\documentclass[runningheads]{llncs}
\usepackage{graphicx}
\usepackage{multirow}
\usepackage{subfigure}
\usepackage{url}
\usepackage{color}
\usepackage{hyperref}

\begin{document}
	\title{Boundary-assisted Region Proposal Networks for Nucleus Segmentation}
	%
	%

	\author{Shengcong Chen\inst{1}\orcidID{0000-0002-8019-9675} \and
		Changxing Ding\inst{1}\orcidID{0000-0001-7232-3181} \and
		Dacheng Tao\inst{2}\orcidID{0000-0001-7225-5449}}
	\authorrunning{S. Chen et al.}
	%
	\institute{School of Electronic and Information Engineering, South China University of Technology, Guangzhou 510641, China\\
	\email{c.shengcong@mail.scut.edu.cn, chxding@scut.edu.cn} \and
	 UBTECH Sydney AI Centre, School of Computer Science, Faculty of Engineering, The University of Sydney, Darlington, NSW 2008, Australia}

	\maketitle

	\begin{abstract}
		
	Nucleus segmentation is an important task in medical image analysis. However, machine learning models cannot perform well because there are large amount of clusters of crowded nuclei. To handle this problem, existing approaches typically resort to sophisticated hand-crafted post-processing strategies; therefore, they are vulnerable to the variation of post-processing hyper-parameters. Accordingly, in this paper, we devise a Boundary-assisted Region Proposal Network (BRP-Net) that achieves robust instance-level nucleus segmentation. First, we propose a novel Task-aware Feature Encoding (TAFE) network that efficiently extracts respective high-quality features for semantic segmentation and instance boundary detection tasks. This is achieved by carefully considering the correlation and differences between the two tasks. Second, coarse nucleus proposals are generated based on the predictions of the above two tasks. Third, these proposals are fed into instance segmentation networks for more accurate prediction. Experimental results demonstrate that the performance of BRP-Net is robust to the variation of post-processing hyper-parameters. Furthermore, BRP-Net achieves state-of-the-art performances on both the Kumar and CPM17 datasets. The code of BRP-Net will be released at \url{https://github.com/csccsccsccsc/brpnet}.

		\keywords{Nucleus segmentation \and Multi-task Learning \and Instance Segmentation.}
		
	\end{abstract}
	
	\section{Introduction}
	\label{secintroduction}
	
	Nucleus segmentation is a crucial task in computational pathology, as it provides rich spatial and morphometric information regarding nuclei. However, automatic nucleus segmentation remains challenging. This is for a number of reasons: first, a large amount of nucleus clusters exist, which results in crowded and overlapping nuclei; second, the boundary of nuclei in out-of-focus images tends to be blurry, which increases the difficulty associated with separating crowded instances; third, both nucleus appearance and shape exhibit dramatic variation, which makes the segmentation task more difficult.
	
	\begin{figure}
		\centering
		\subfigure[]
		{
			\includegraphics[width=0.2\textwidth]{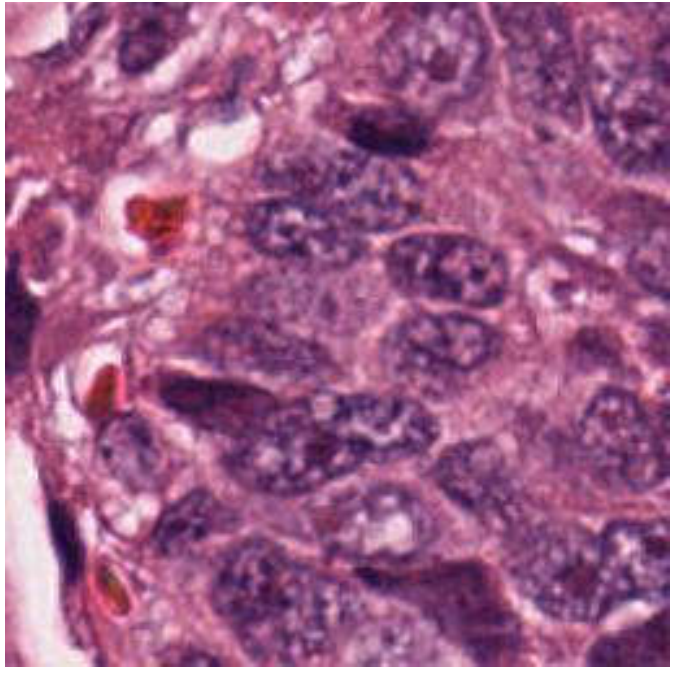}
			\label{fig1:img}
		}
		\quad
		\quad
		\quad
		\subfigure[]
		{
			\includegraphics[width=0.2\textwidth]{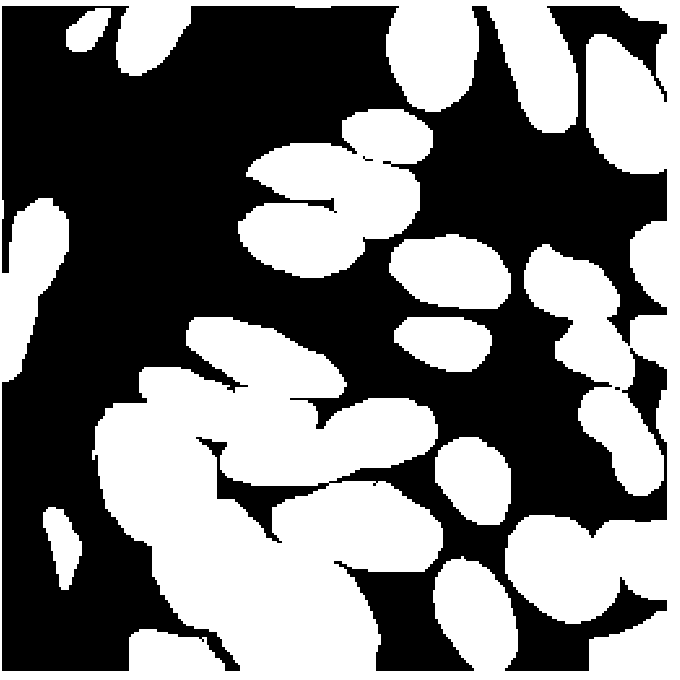}
			\label{fig1:seg}
		}
		\quad
		\quad
		\quad
		\subfigure[]
		{
			\includegraphics[width=0.2\textwidth]{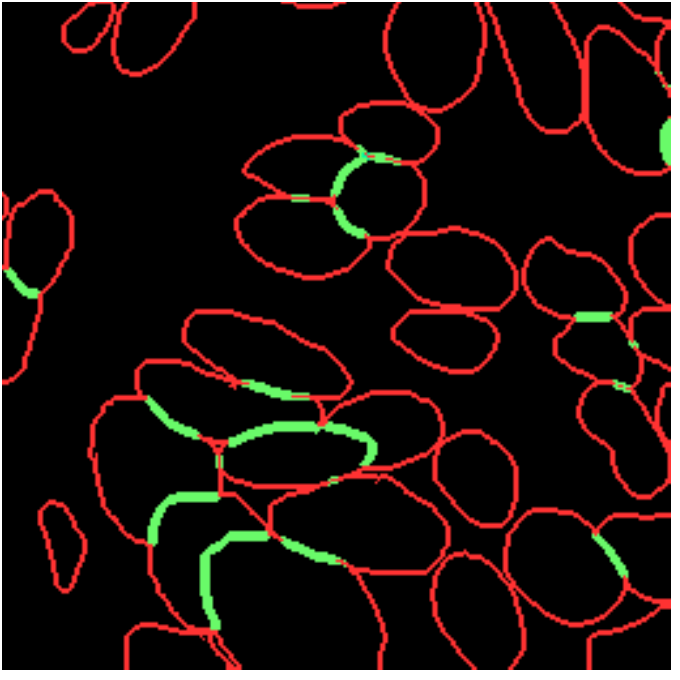}
			\label{fig1:bnd}
		}
		\caption
		{
			An example that illustrates the essential difference between the semantic segmentation task and the instance boundary detection task. (a) original image; (b) ground-truth for semantic segmentation; (c) ground-truth for nucleus boundary. The boundaries that separates two overlapping instances, i.e. pixels colored in green in (c), cannot be directly inferred from the semantic segmentation results in (b).
		}
		\label{fig1}
	\end{figure}
	
	Many approaches to nucleus segmentation have been proposed. One popular scheme is based on the use of boundary detection \cite{dcan,bes-net,cia-net}. These approaches subtract instance boundaries from semantic segmentation results and then employ complex post-processing rules to obtain specific instances. In order to obtain the instance boundaries, DCAN \cite{dcan} adopted two decoders for U-Net, one for semantic segmentation and another for instance boundary detection. No interactions take place between the two decoders. To make use of their correlation, BES-Net \cite{bes-net} and CIA-Net \cite{cia-net} further introduced uni-directional and bi-directional information transmission, respectively, which means one decoder obtains extra features from the other one. There are two key downsides of the above approaches. First, as they adopt a shared encoder for both tasks, they consequently underestimate the essential differences between tasks in feature learning; for example, the boundaries in Fig.~\ref{fig1:bnd} that separate two overlapping instances cannot be directly inferred from the semantic segmentation results in Fig.~\ref{fig1:seg}. Second, because these approaches adopt complex post-processing rules, their performance is sensitive to the variation of post-processing hyper-parameters.
	
	Another popular strategy used to separate crowded instances is the distance-based approach \cite{dist,hover-net}. For example, DIST \cite{dist} predicted the distance between each foreground pixel and its nearest background pixel, while HoVer-Net \cite{hover-net} enriched prediction by considering distances in both the horizontal and vertical directions. Subsequently, these works apply the watershed algorithm to the predicted distance maps to obtain instances. However, one downside of this approach is that the watershed algorithm may be sensitive to the noise in the distance maps. Finally, clustering-based methods predict the spatial location of the associated instance for each foreground pixel \cite{spa-net}. These instances are separated by clustering the predicted location coordinates.
	
	\begin{figure}
		\centering
		\includegraphics[width=0.7\textwidth]{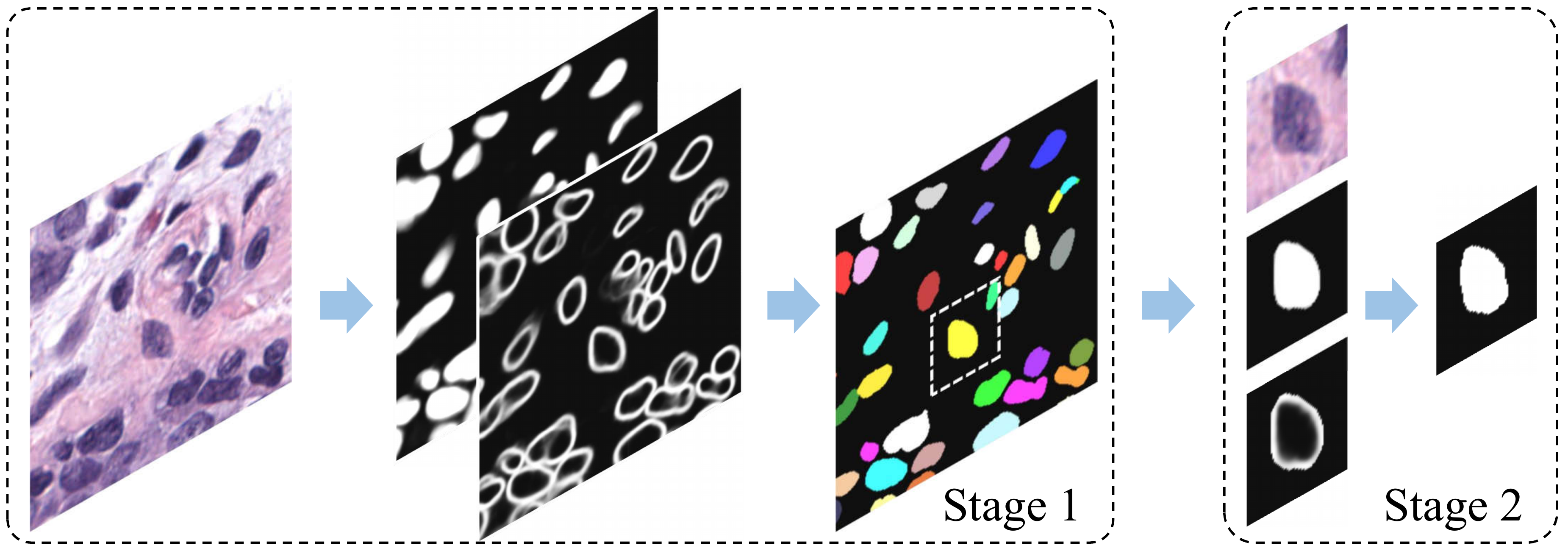}
		\caption{Overview of BRP-Net. BRP-Net comprises two stages: one stage to obtain instance proposals and another for proposal-wise segmentation.}\label{fig2}
	\end{figure}
	
	In this paper, we propose a novel framework for nucleus segmentation, referred to as Boundary-assisted Region Proposal Network (BRP-Net). Similar to Mask R-CNN \cite{maskrcnn}, BRP-Net comprises two stages: one stage to obtain instance proposals and another for proposal-wise segmentation. In the first stage, we implement the boundary detection-based scheme to obtain instance proposals. This can be contrasted with Mask R-CNN \cite{maskrcnn}, which predicts rectangular proposals directly from feature maps. As was demonstrated in \cite{shape-aware-rcnn}, crowded instances result in bounding boxes with significant overlap; this means a single bounding box can be associated with multiple instances, consequently affecting the optimization quality of the network. Moreover, we further propose the Task-aware Feature Encoding (TAFE) network, which efficiently extracts high-quality features for semantic segmentation and instance boundary detection tasks. TAFE aids BRP-Net in robustly obtaining instance proposals. The second stage refines the segmentation result for each proposal, which enables BRP-Net to be robust to the variation of post-processing hyper-parameters in TAFE. Extensive experiments are conducted on two publicly available nucleus segmentation datasets, from which we can conclude that BRP-Net consistently achieves state-of-the-art performance on both datasets.

	\section{Method}
	\label{secmethod}
	
	The overall framework of BRP-Net is presented in Fig.~\ref{fig2}. This framework includes two stages: one for obtaining instance proposals and another for proposal-wise segmentation. The first stage adopts a similar pipeline to CIA-Net \cite{cia-net}, and the second one aims to refine the segmentation results of the first stage in a proposal-wise manner.
	
	\subsection{Region Proposal Generation}
	\begin{figure}
		\centering
		\includegraphics[width=0.9\textwidth]{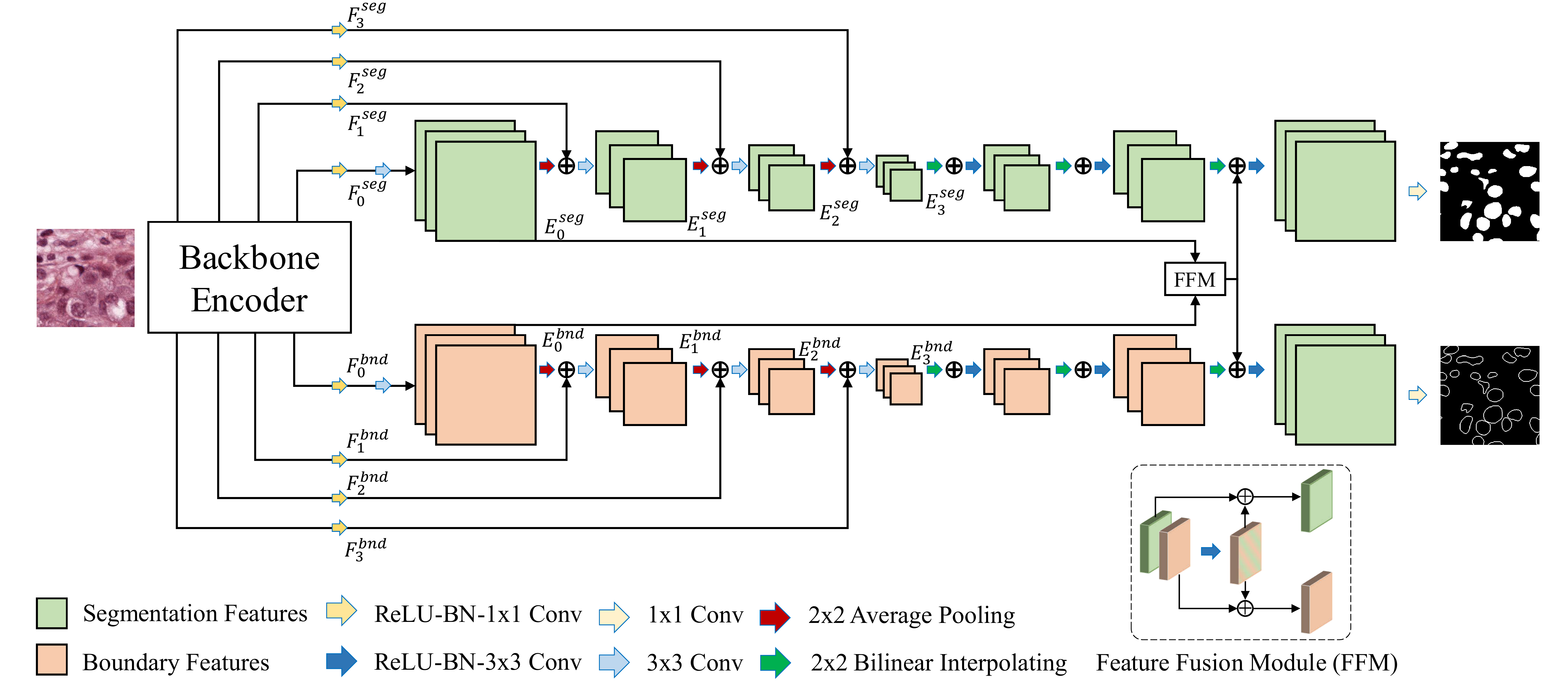}
		\caption{
			Architecture details of TAFE. The number of channels in $F_{i}$ is set to 256 consistently. Feature maps produced by both encoders are fused in FFMs to make use of their correlation. For simplicity, only one FFM is shown and the other two FFMs are ignored in this figure. (Best viewed in color).
		}
		\label{fig3}
	\end{figure}
	We adopt a boundary detection-based scheme to obtain high-quality region proposals. Following the post-processing rules outlined in \cite{dcan,cia-net}, instance boundaries are subtracted from the predictions of semantic segmentation. Subsequently, connected component analysis is applied to produce instance proposals. Extant approaches have integrated semantic segmentation and instance boundary detection tasks into one model \cite{dcan,bes-net,cia-net}; however, as they adopt a shared encoder for both tasks, they may underestimate their essential differences regarding feature learning, as is analyzed in Sec.~\ref{secintroduction}. One intuitive solution would be adopting independent encoders for the two tasks. However, this strategy increases the model complexity and also completely ignores their correlation. Accordingly, we propose a novel Task-aware Feature Encoding (TAFE) network capable of efficiently extracting high-quality features for each of these tasks.

	Fig.~\ref{fig3} presents the architecture of TAFE. First, nucleus images are fed into a single backbone encoder to extract feature maps that are \{1, 1/2, 1/4, 1/8\} of the original image size. The structure of the backbone encoder is provided in the supplementary file. Subsequently, each of them is passed through one unshared $1\times1$ convolutional layer to obtain $F^{seg}_{i}$ and $F^{bnd}_{i}$. $F^{seg}_{i}$ and $F^{bnd}_{i}$ are fed into Task-specific Encoders (TSE), which are designed for the semantic segmentation and instance boundary detection tasks, respectively. In each encoder, feature maps after down-sampling are merged with an $F_{i}$ of the same size via element-wise summation. The merged features are then passed through one $3\times3$ convolutional layer to generate $E_{i}$. Similar to CIA-Net \cite{cia-net}, deep supervision is applied and the auxiliary classifiers take $E_{i}$ as inputs. Moreover, inspired by the Information Aggregation Modules \cite{cia-net}, we propose the light-weight Feature Fusion Modules (FFMs), which is based on residual learning to aggregate information in $E^{seg}_{i}$ and $E^{bnd}_{i}$. In the experimentation section, we demonstrate the superiority of FFMs. FFMs are helpful for making use of the correlation as well as reserving the differences between both tasks. Outputs of each FFM are fed into two shallow decoders via element-wise summation. The two decoders are used for the semantic segmentation task and the instance boundary detection task. Each decoder contains three BN-ReLU-Conv layers.
	
	\subsection{Proposal-wise Segmentation}
	\begin{figure}
		\centering
		\includegraphics[width=0.9\textwidth]{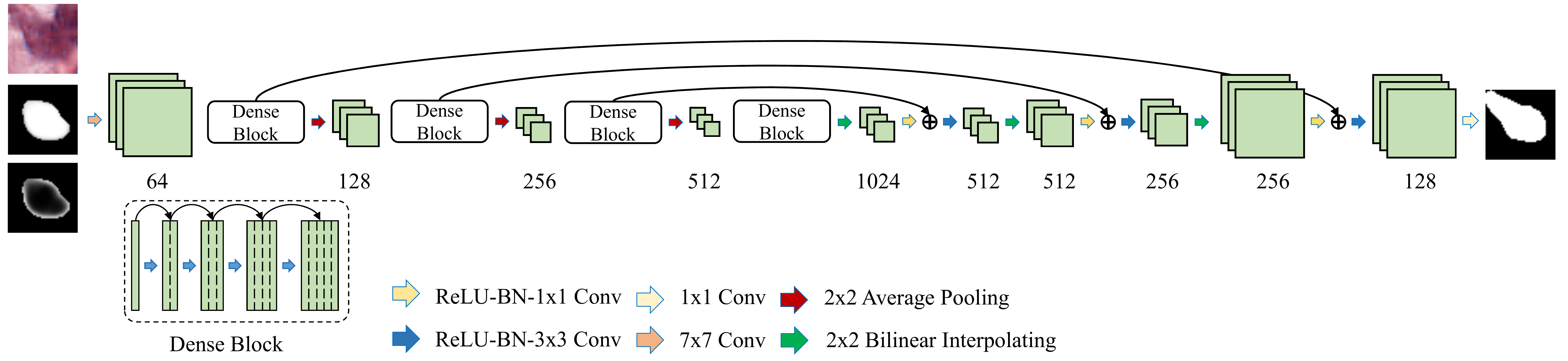}
		\caption{The two networks in the proposal-wise segmentation stage adopt the same architecture. Each layer in the network includes one dense block that consists of four $3{\times}3$ convolutional layers. Growth rates of the four dense blocks are set to 16, 32, 64, and 128, respectively. The number below each group of feature maps denotes the number of channels. (Best viewed in color).}\label{fig4}
	\end{figure}
	The first stage of BRP-Net, i.e. TAFE, adopts hand-crafted post-processing rules to obtain instance proposals. Accordingly, the quality of proposals is affected by post-processing hyper-parameters. To address this problem, we propose a second stage for BRP-Net to facilitate more robust segmentation.
	
	We crop one square patch containing each proposal with a minimal margin of 12 pixels on each side. Because the patches vary dramatically in size, we group them into small and large patches with a threshold of $S_S$ according to their length. Then, small and large patches are resized to $S_S{\times}S_S$ and $S_L{\times}S_L$, respectively. Finally, we train one network for the small and another for the large patches. These two networks have the same architecture, the details of which are illustrated in Fig.~\ref{fig4}. Inputs to the model include the patch, and the probability maps that are predicted by the semantic segmentation and boundary detection tasks in the first stage. To relieve the influence of background, elements in the probability maps that fall outside of the dilated proposal are set to zero. The dilation rate is set to 2 pixels.
	
	During training, each proposal is matched to a ground-truth instance depending on their Intersection over Union (IoU). For proposals with an IoU larger than $\tau$, their label maps are set with reference to the matched ground-truth instance; otherwise, the proposals are considered to be false-positive predictions. Therefore, all elements in their label maps are set to zero (denoting background).
	\subsection{Inference}
	During the inference process, nucleus images are fed into BRP-Net. Semantic segmentation and instance boundary detection results are produced by TAFE. Then, post-processing operations in \cite{dcan,cia-net} are implemented to obtain instance proposals. Finally, patches containing these proposals are extracted and respectively fed into proposal-wise segmentation networks for robust instance segmentation.

	\section{Experiments}
	We conduct experiments on two publicly available datasets. The first is a multi-organ nucleus dataset \cite{kumar,kumar2}, referred to as Kumar, which contains 30 Hematoxylin and Eosin ($H \& E$) stained images with resolution of $1000\times1000$. They are divided into a training set of 16 images and a testing set of 14 images according to the same protocol used in previous works \cite{kumar,cia-net,hover-net,spa-net}. In the testing set, 8 images are from 4 organs in the training set (seen organ), and the remained 6 images are from 3 organs that do not appear in the training set (unseen organ). The second dataset is Computational Precision Medicine Dataset (CPM17) \cite{cpm17}, which contains 32 images for training and 32 images for testing.
	
	Evaluation metrics for the two datasets are different. In the Kumar dataset, the main metric is the Average Jaccard Index (AJI) \cite{kumar}. We also report the F1-Score to measure the instance detection performance. In CPM17, we use the same metrics as used in \cite{cpm17}, i.e. the DICE coefficient (DICE 1) and Ensemble Dice (DICE 2). DICE 1 measures the overall overlap between the predictions and the ground truth, and DICE 2 measures the average overlap between the predictions and their matched ground truth instances. Besides, in order to better compare with one state-of-the-art work \cite{hover-net}, we also report AJI in the experiments.
	
	\subsection{Implementation Details}
	We first perform stain normalization \cite{stainnorm} to reduce the color differences between the stained images. In the next step, we normalize each image by subtracting the mean and dividing by the standard deviation of the training set. Training data are augmented by random cropping, flipping, color jittering, blurring and elastic transformation. We crop images to a size of $256\times256$ pixels before using them as the input of BRP-Net.
	
	In a similar way to CIA-Net \cite{cia-net}, we adopt DenseNet \cite{densenet} as TAFE’s backbone encoder and initialize its parameters using a single pretrained model\setcounter{footnote}{0}
	\renewcommand{\thefootnote}{\fnsymbol{footnote}}\footnote{The pretrained model can be downloaded from \url{https://download.pytorch.org/models/densenet121-a639ec97.pth}}. We also adopt both the Smooth Truncated Loss \cite{cia-net} and Soft Dice Loss \cite{vnet} for the optimization of both tasks in TAFE. Weight of the Soft Dice Loss is set to 0.5. We use the AdamW \cite{adamw} optimizer for training. The number of training epochs is set to 600. The learning rate is initially set to 0.0003, and decreases according to the cosine annealing schedule \cite{adamw}. The learning rate decreases to zero in 40 epochs and is then reset. At each restart, the new start learning rate is set to be one half of the previous rate, while the new period lasts for twice as long as the previous one.
	
	Finally, $S_S$ and $S_L$ are set to be 48 and 176 pixels, respectively. Training settings for the proposal-wise segmentation networks are similar to those of TAFE. But we use Focal Loss \cite{focalloss} for optimization and the training lasts for only ten epochs. The learning rate is set to 0.0003 initially, and decreases according to the cosine annealing schedule without restart.
	
	\subsection {Ablation Study}
	\subsubsection {Effectiveness of TAFE}
	\begin{table}
	\centering
	\caption{Performance comparisons between the baseline, baseline+FFMs, and TAFE.}
	\renewcommand\tabcolsep{7.5pt}
	\begin{tabular}{c|c|c|c|c|c|c}
		\hline
		\multirow{2}{*}{Network} & \multicolumn{3}{c|}{AJI (\%)} & \multicolumn{3}{c}{F1-Score (\%)} \\
		\cline{2-7}
		&seen &unseen &all &seen &unseen &all\\
		\hline
		Baseline & 61.15 & 62.58 & 61.76 & 82.99 & 84.08 & 83.46 \\
		\hline
		Baseline+FFMs & 61.41 & 63.39 & 62.26 & 82.35 & \textbf{84.90} & 83.44 \\
		\hline
		TAFE & \textbf{61.96} & \textbf{63.84} & \textbf{62.77} & 82.81 & 84.34 & 83.47 \\
		\hline
	\end{tabular}
	\label{tab1}
	\end{table}
	\begin{figure}
	\centering
	\subfigure[] {
		\label{fig:radii}
		\includegraphics[width=0.48\textwidth]{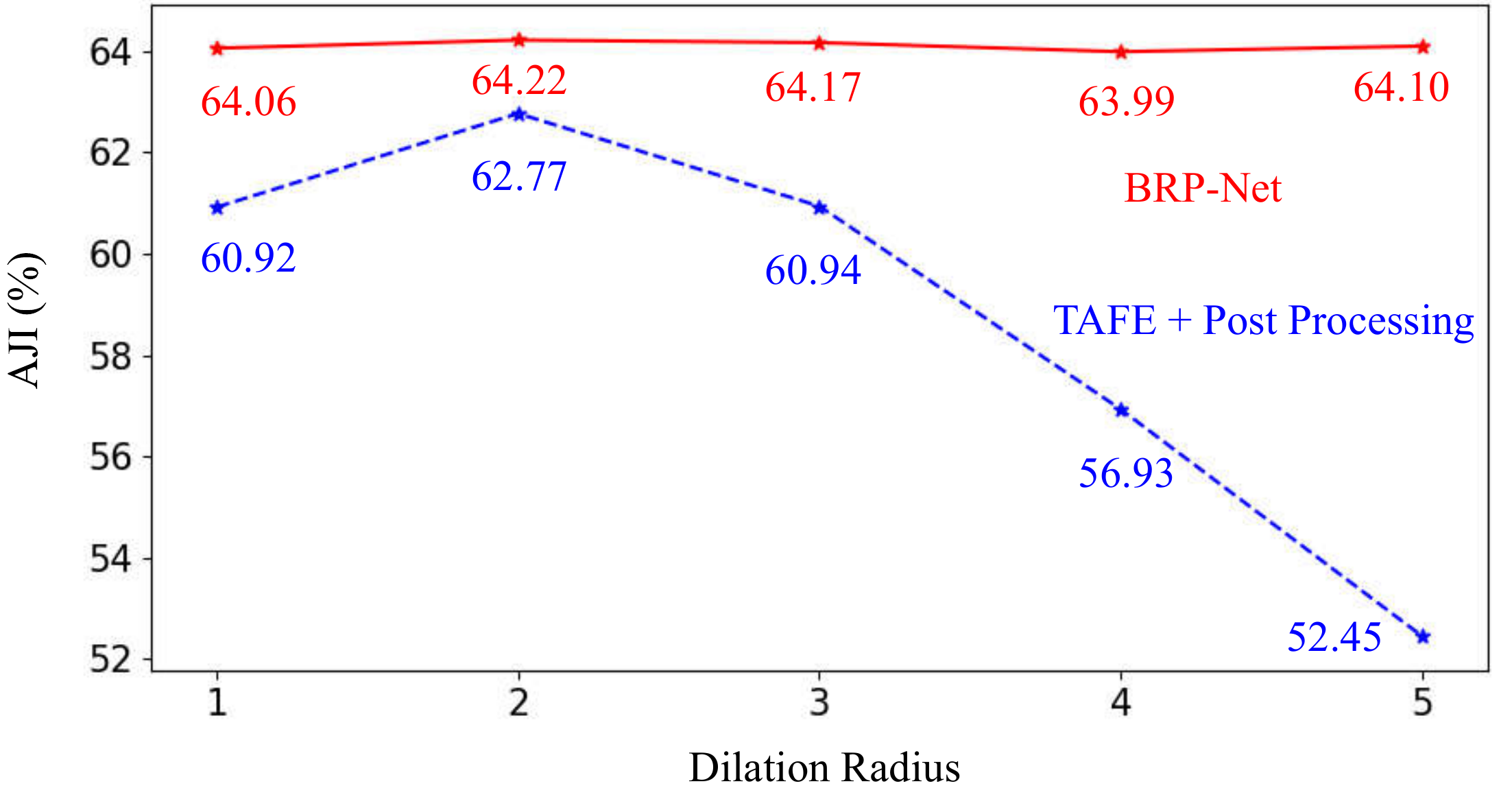}}
	\subfigure[] {
		\label{fig:train}
		\includegraphics[width=0.48\textwidth]{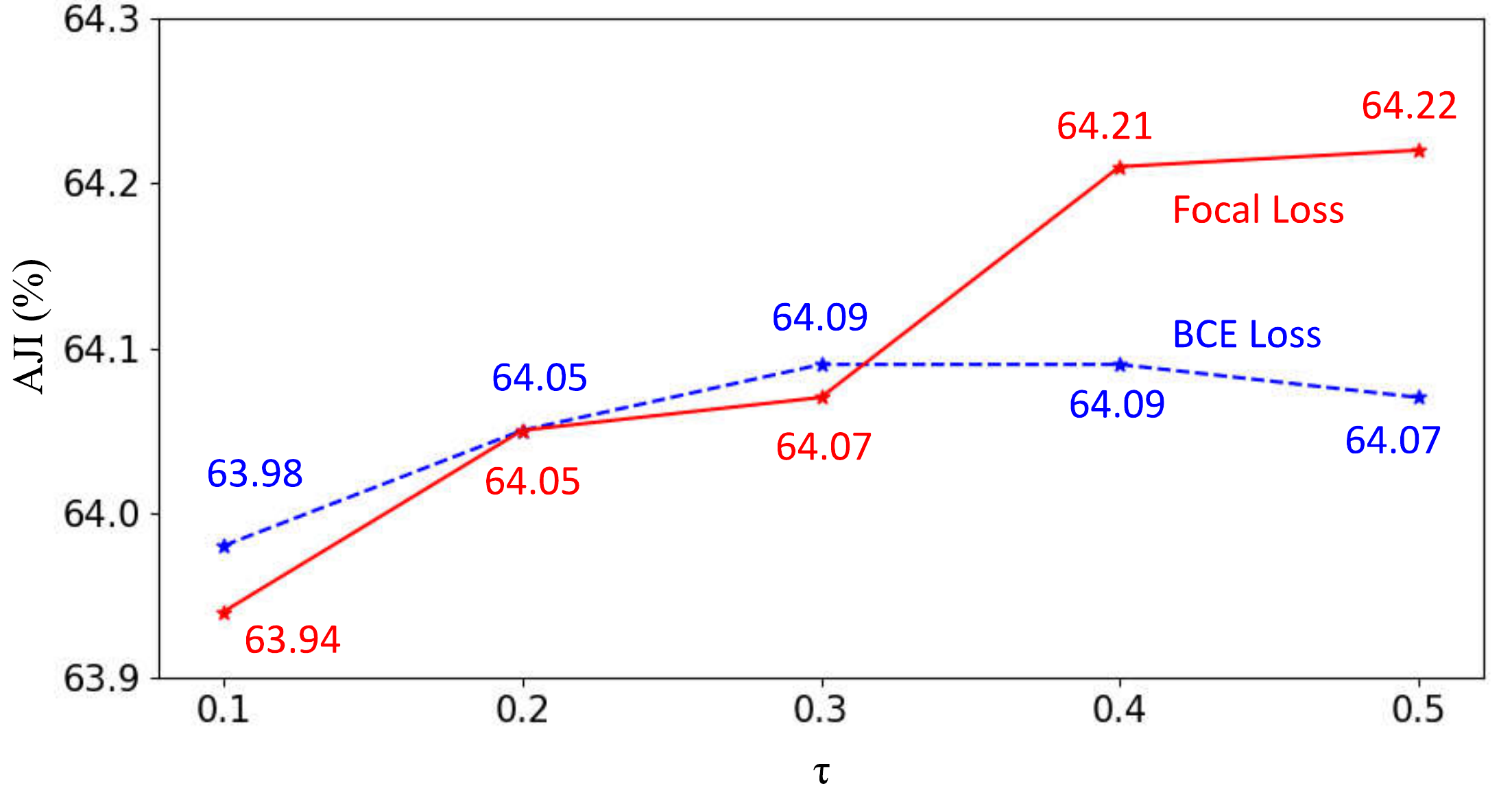}}
	\caption{Evaluation on different settings for BRP-Net. (a) The influence of different dilation radii in the post-processing step of TAFE. (b) The choice of IoU thresholds $\tau$ and different loss functions for the second stage of BRP-Net.}\label{fig5}
	\end{figure}
	We compare the performance of TAFE with a baseline network that is similar to existing boundary detection-based methods \cite{cia-net}. In brief, it shares encoder for the semantic segmentation and instance boundary detection tasks. The two tasks still own respective decoders equipped with IAMs. For fair comparison, the baseline has the same number of parameters as TAFE. Table~\ref{tab1} presents the performance of TAFE and baseline. We also equip IAMs with the same residual learning scheme as FFMs and report the performance of baseline again, which is referred to as `baseline+FFMs' in the table. Architecture details of both the baseline and ‘baseline+FFMs’ are provided in the supplementary file. It can be seen from our results that TAFE achieves higher AJI performance on both the seen and unseen organ datasets. It is also clear that the residual learning scheme in FFMs is helpful. This may be because this scheme better highlights the differences between the two tasks, as illustrated in Fig.~\ref{fig3}. The comparison justifies the effectiveness of TAFE and the FFM modules.
	
	\subsubsection {Evaluation on Post-Processing Settings in TAFE}
	Existing boundary detection-based methods are sensitive to the post-processing hyper-parameters, particularly the dilation radius for recovering the subtracted instance boundaries \cite{dcan,cia-net}. We conduct experiments to evaluate the influence of different dilation radii on both TAFE and the entire BRP-Net pipeline. Results are presented in Fig.~\ref{fig:radii}. It can be found that due to the proposal-wise segmentation stage, BRP-Net is highly robust to the value of dilation radius. By contrast, the performance of single-stage method is less stable.

	\subsubsection {Evaluations on Settings for Proposal-wise Segmentation}
	We evaluate the influence of IoU thresholds $\tau$ and different loss functions on the second stage of BRP-Net. Experimental results are presented in Fig.~\ref{fig:train}. It is shown that the performance of BRP-Net is generally robust to the value of $\tau$, as well as that focal loss \cite{focalloss} slightly outperforms cross-entropy loss. According to the evaluation results, we select focal loss for training and set $\tau$ as 0.5 for the second stage.
	\begin{table}
	\centering
	\renewcommand\tabcolsep{7.5pt}
	\caption{Quantitative comparisons between BRP-Net and existing methods.}
	\subtable[Comparisons on the Kumar database \cite{kumar}.]{
		\begin{tabular}{c|c|c|c|c|c|c}
			\hline
			\multirow{2}{*}{Network} & \multicolumn{3}{c|}{AJI (\%)} & \multicolumn{3}{c}{F1-Score (\%)} \\
			\cline{2-7}
			&seen &unseen &all &seen &unseen &all\\
			\hline
			CNN3 \cite{kumar} & 51.54 & 49.89 & 50.83 & 82.26 & 83.22 & 82.67 \\
			\hline
			DIST \cite{dist} & 55.91 & 56.01 & 55.95 & - & - & -\\
			\hline
			Mask R-CNN \cite{maskrcnn} & 59.78 & 55.31 & 57.86 & 81.07 & 82.91 & 81.86\\
			\hline
			CIA-Net \cite{cia-net} & 61.29 & 63.06 & 62.05 & 82.44 & 84.58 & 83.36 \\
			\hline
			HoVer-Net \cite{hover-net} & - & - & 61.80 & - & - & - \\
			\hline
			Spa-Net \cite{spa-net} & 62.39 & 63.40 & 62.82 & 82.81 & 84.51 & 83.53 \\
			\hline
			BRP-Net (ours) & \textbf{63.07} & \textbf{65.75} & \textbf{64.22} & \textbf{83.46} & \textbf{85.26} & \textbf{84.23}\\
			\hline
		\end{tabular}
		\label{tab:kumar}
	}
	\subtable[Comparisons on CPM17 database \cite{cpm17}]{
		\begin{tabular}{c|c|c|c}
			\hline 
			Network & Dice 1 (\%) & Dice 2 (\%) & AJI (\%) \\ 
			\hline
			DRAN \cite{cpm17} & 86.2 & 70.3 & 68.3 \\
			\hline
			HoVer-Net \cite{hover-net} & 86.9 & - & 70.5 \\
			\hline
			Micro-Net \cite{micronet} & 85.7 & \textbf{79.6} & - \\
			\hline
			BRP-Net (ours) & \textbf{87.7} & 79.5 & \textbf{73.1} \\
			\hline
		\end{tabular}
		\label{tab:cpm17}
	}
	\label{tab2}
	\end{table}

	\subsection {Comparisons with State-of-the-art Methods}
	Comparisons between BRP-Net and state-of-the-art methods on the Kumar database are reported in Table~\ref{tab:kumar}. It can be seen that BRP-Net achieves both the highest AJI and the highest F1-Score among all the methods. In particular, BRP-Net outperforms the previous best method, i.e. SPA-Net, by 0.68\%, 2.35\%, and 1.40\% on the seen organ, unseen organ, and all testing data, respectively. We also provide qualitative comparisons in the supplementary file.

	We further conduct comparisons on CPM17 database \cite{cpm17} and summarize the results in Table~\ref{tab:cpm17}. From the results, we can see that BRP-Net continues to achieves state-of-the-art performance. Its performance in Dice 1 and AJI outperforms existing approaches by 0.8\% and 2.6\%, respectively. The above comparisons demonstrate the effectiveness of BRP-Net.

	\section {Conclusion}
	In this paper, we propose the Boundary-assisted Region Proposal Network (BRP-Net) for nucleus segmentation. BRP-Net contains one stage designed for obtaining instance proposals and a second stage for proposal-wise segmentation. To separate crowded nuclei, we adopt a boundary detection-based scheme for the first stage. We further propose a novel Task-specific Feature Encoding network with Feature Fusion Modules to achieve this goal. The second stage is further introduced to segment proposals of various size, and enables BRP-Net to be robust to the variation of post-processing hyper-parameters in the first stage. Finally, BRP-Net achieves strong performance on both the Kumar and CPM17 datasets.
	
	\section *{Acknowledgement}
	Changxing Ding is supported by NSF of China under Grant 61702193 and U1801262, the Science and Technology Program of Guangzhou under Grant 201804010272, the Program for Guangdong Introducing Innovative and Entrepreneurial Teams under Grant 2017ZT07X183, and the Fundamental Research Funds for the Central Universities of China under Grant 2019JQ01. Dacheng Tao is supported by Australian Research Council Project FL-170100117.
	~\\

	%
	%
	%
	%

	\newpage
	\title{Supplementary Material}
	\author{}
	\institute{}
	\maketitle
	\setcounter{section}{0}
	\section{Architecture Details of the Backbone Encoder}
	\begin{figure}
		\centering
		\includegraphics[width=0.93\textwidth]{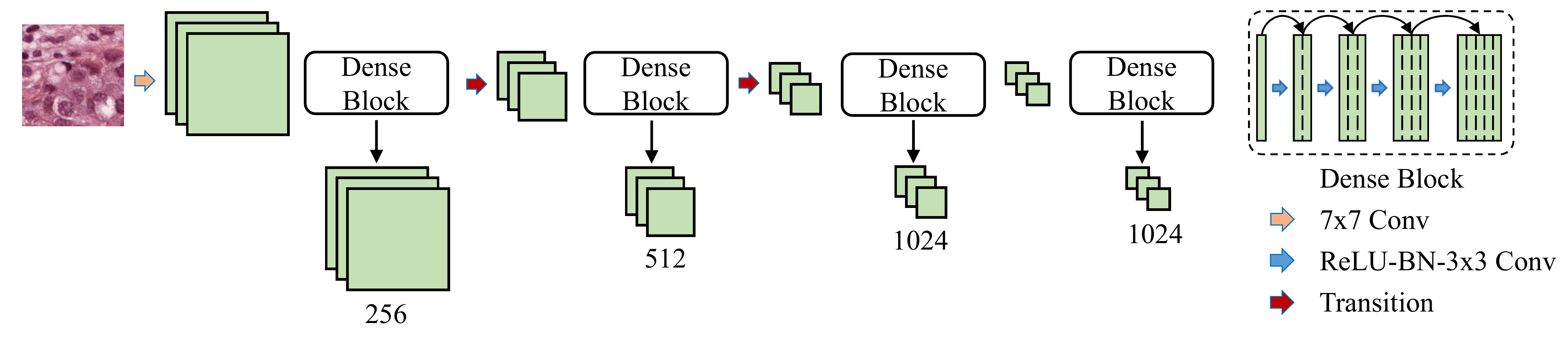}
		\caption{Architecture details of the backbone encoder in Sec. 2.1 of the main paper. We adopt DenseNet121 as the backbone. The four dense blocks contain 6, 12, 18 and 24 convolutional layers, respectively, and the growth rate is set to 32. The stride of the first $7\times7$ convolutional layer is set to 1 and the first max-pooling layer is removed. Therefore, sizes of the obtained feature maps from the four dense blocks are 1, 1/2, 1/4 and 1/8 of the input image size, respectively. (Best viewed in color).}
	\end{figure}
	
	\section{Architecture Details of the Baseline}
	\begin{figure}
		\centering
		\includegraphics[width=0.825\textwidth]{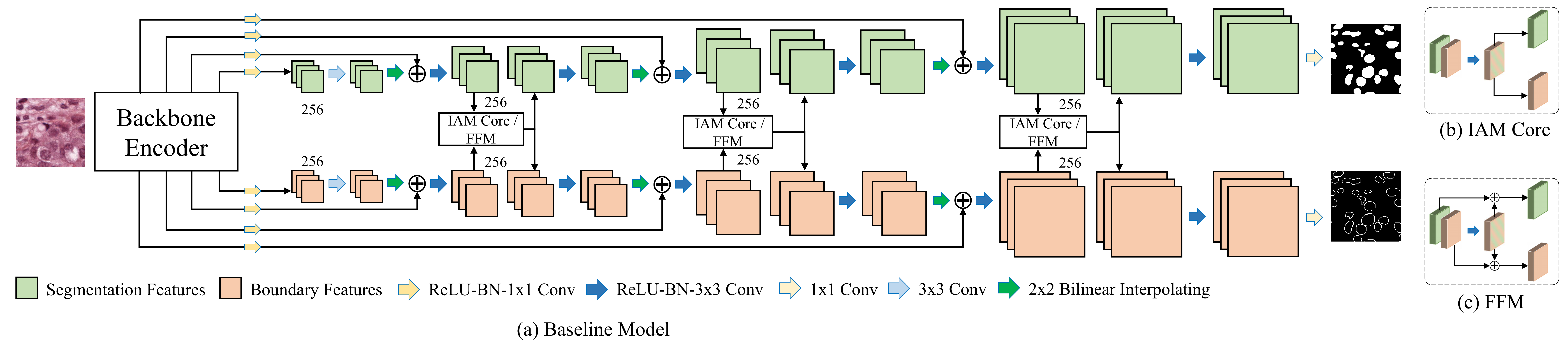}
		\caption{Architecture details of the baseline and `baseline+FFMs' in Table 1 of the main paper. Their difference is that the former adopts Information Aggregation Modules (IAMs), while the latter equips IAMs with the same residual learning scheme as Feature Fusion Modules (FFMs). `IAM Core' refers to the core components for information aggregation in IAM, as illustrated in (b). The baseline and `baseline+FFMs' has the same number of parameters as TAFE. (Best viewed in color).}
	\end{figure}
	\newpage
	
	\section{Qualitative Comparisons}
	\begin{figure}
		\centering
		\includegraphics[width=0.978\textwidth]{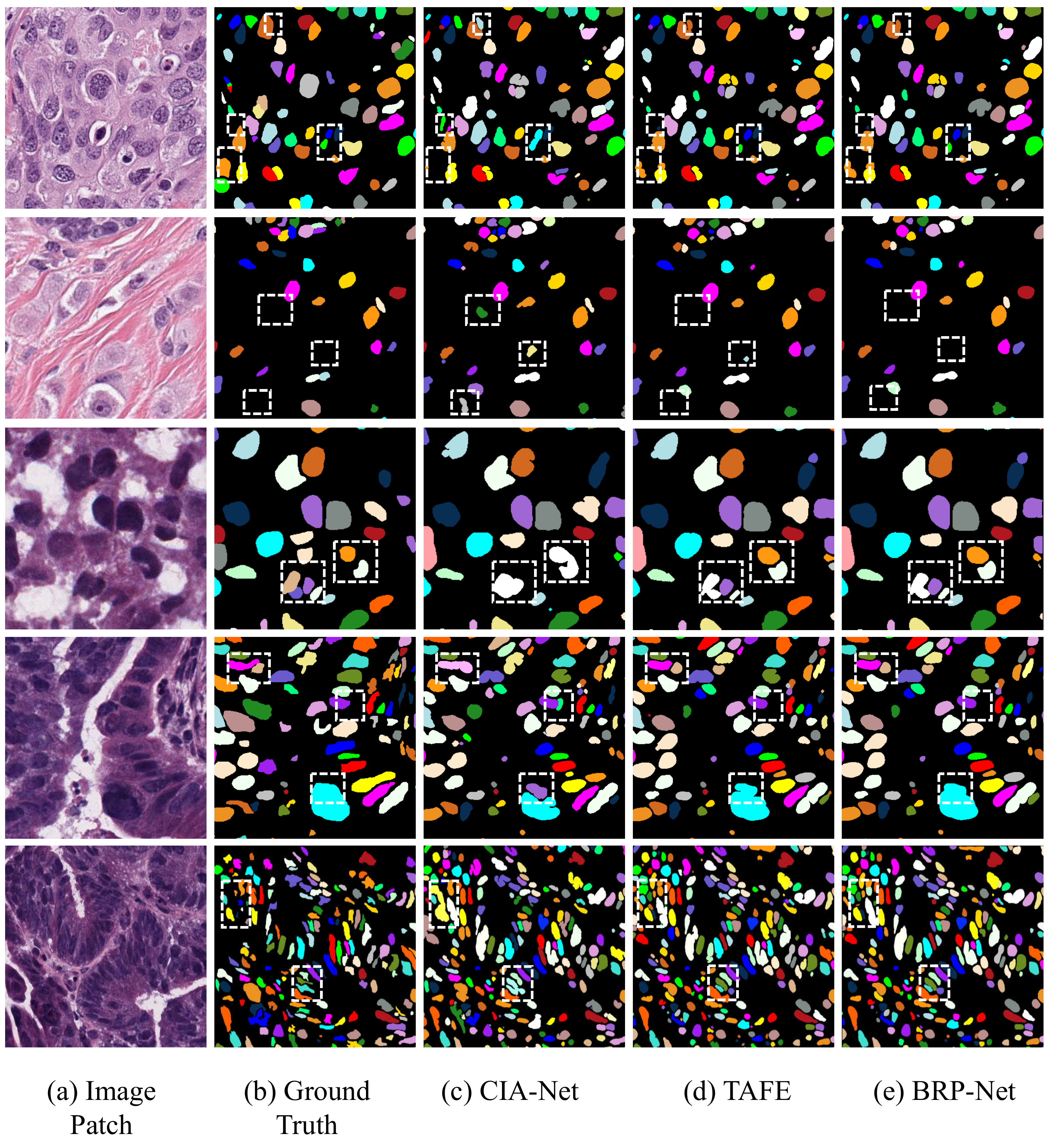}
		\caption{Qualitative comparisons between different models. From left to right in each row: the original image, ground truth segmentations, the predictions by CIA-Net [3], TAFE, and BRP-Net. (Best viewed in color).}
	\end{figure}

\end{document}